\documentclass{new_tlp}
\usepackage{mathptmx}
\usepackage{acronym}

\newacro{ONoC}{Optical Network on Chip}
\newacroplural{ONoC}{Optical Networks on Chip}
\newacro{WRONoC}{Wavelength-Routed Optical Network on Chip}
\newacroplural{WRONoC}{Wavelength-Routed Optical Networks on Chip}
\newacro{PS}{Photonic Switch}
\newacroplural{PS}{Photonic Switches}
\newacro{TWC}{Tunable Wavelength Converter}
\newacro{ASP}{Answer Set Programming}
\newacro{IC}{Integrity Constraint}
\newacro{CLP}{Constraint Logic Programming}
\newacro{CLP(FD)}{Constraint Logic Programming on Finite Domains}
\newacro{CSP}{Constraint Satisfaction Problem}
\newacro{COP}{Constraint Optimization Problem}
\newacro{MILP}{Mixed-Integer Linear Programming}

\usepackage{listings}
\usepackage[draft]{todonotes}
\usepackage{amsmath}

 \usepackage[usenames,dvipsnames]{pstricks}
 \usepackage{epsfig}
 \usepackage{pst-grad} 
 \usepackage{pst-plot} 
 \usepackage[space]{grffile} 
 \usepackage{etoolbox} 
 \makeatletter 
 \patchcmd\Gread@eps{\@inputcheck#1 }{\@inputcheck"#1"\relax}{}{}
 \makeatother

\newcommand{\arxiv}[1]{#1}   

\usepackage{macro}

\newcommand{\Nlambdas}{{\ensuremath{n_{\lambda}}}}
\newcommand{\Ndisp}{{\ensuremath{n_R}}}
\newcommand{\Sender}{{\ensuremath{Se}}}
\newcommand{\Receiver}{{\ensuremath{Re}}}
\newcommand{\pred}[2]{{\tt #1}$/#2$}
\newcommand{\lamb}{{\ensuremath{\lambda}}}
\newcommand{\SetPossRadius}{\bR} 
\newcommand{\SetPossLambda}{{\ensuremath{\Lambda}}} 
\newcommand{\Dist}{{\ensuremath{Dist}}} 
\newcommand{\MaxLambdasPerRadius}{{\ensuremath{Max_{\sharp\lambda}}}} 
\newcommand{\MLPR}{\MaxLambdasPerRadius}

\newcommand{\alldifferent}{{\ensuremath{\mbox{\textit{alldifferent}}}}}
\newcommand{\cumulative}{{\ensuremath{\mbox{\textit{cumulative}}}}}
\newcommand{\eclipse}{ECL$^i$PS$^e$}

\newcommand{\Mat}{{\ensuremath{M}}}  

  \title[Logic Programming approaches for routing fault-free and maximally-parallel WRONoCs]
        {Logic Programming approaches for routing fault-free and maximally-parallel Wavelength Routed Optical Networks on Chip
        (Application paper)}

  \author[M. Gavanelli et al.]
         {MARCO GAVANELLI, MADDALENA NONATO\\
         Department of Engineering, Ferrara University, Ferrara, Italy\\
         \email{[marco.gavanelli,maddalena.nonato]@unife.it}
         \and
         ANDREA PEANO \\
         QuanTek, Bologna, Italy\\
         \email{andrea.peano@quantek.it} \and
         DAVIDE BERTOZZI \\
         Department of Engineering, Ferrara University, Ferrara, Italy\\
         \email{davide.bertozzi@unife.it}
  }

\jdate{March 2003}
\pubyear{2003}
\pagerange{\pageref{firstpage}--\pageref{lastpage}}
\doi{S1471068401001193}

\begin{document}

\lstset{language=Prolog,
    basicstyle=\tt,         
    commentstyle=\textcolor[rgb]{0.00,0.59,0.00},       
    showstringspaces=false, 
    keywordstyle = \bf,
    morekeywords = {maximize,min,count},
    mathescape=true         
 }

\label{firstpage}

\maketitle

  \begin{abstract}
One promising trend in digital system integration consists of boosting on-chip communication performance by means of silicon photonics, thus materializing the so-called Optical Networks-on-Chip (ONoCs). 
Among them, wavelength routing can be used to route a signal to destination by univocally associating a routing path to the wavelength of the optical carrier. Such wavelengths should be chosen so to minimize interferences among optical channels and to avoid routing faults. 
As a result, 
physical parameter selection of such networks 
requires the solution of complex constrained optimization problems.
In previous work, published in the proceedings of the International Conference on Computer-Aided Design,
we proposed and solved the problem of computing the maximum parallelism obtainable in the communication between any two endpoints
while avoiding 
misrouting of optical signals.
The underlying technology, only quickly mentioned in that paper, is Answer Set Programming (ASP).
In this work, we detail the ASP approach we used to solve such problem.

Another important design issue is to select the 
wavelengths of optical carriers
such that they are spread across the available spectrum,
in order to reduce the likelihood that, due to imperfections in the 
manufacturing
process, 
unintended routing faults arise.
We show how to address such problem in Constraint Logic Programming on Finite Domains (CLP(FD)).

\arxiv{This paper is under consideration for possible publication on Theory and Practice of Logic Programming}
  \end{abstract}


\section{Introduction}

Since photons move faster than electrons in the matter, and they dissipate lower power in the process,
the new technology of 
silicon photonics
is a great promise for 
small-scale ICT. 
It promises to provide unmatched communication bandwidth and reduced latencies with low energy-per-bit overhead. 
%
In recent years, 
remarkable advances of CMOS-compatible silicon photonic components have made it possible to conceive optical links and switching fabrics for performance- and power- efficient communication on the silicon chip.
One proposal is to have silicon photonics-enabled on-chip interconnection networks implemented entirely in optics and using all-to-all conflict-free communication (leveraging the principle of wavelength-selective routing). 

Wavelength-routed optical networks univocally associate the wavelength of an optical signal with a specific lightpath across the optical transport medium. They started to gain momentum in the domain of wide-area networks 
when it became clear that 
the electronics inside the optical network nodes were becoming the data transmission bottleneck \cite{4542876}. Consequently, lightpaths in wavelength-routed networks were used  to provide all-optical transmission between the source and the destination nodes \cite{153361}. This way, no optical-to-electrical-to-optical conversion and data processing were required at any intermediate node. 

The recent advances of silicon photonics 
have raised a strong interest in using optical networks for on-chip communication (\acp{ONoC}). In this context, wavelength routing has been  proposed as a way of relieving the latency and power overhead of electrically-assisted \acp{ONoC} to resolve optical contention. In fact, \acp{WRONoC} are appealing as all-optical solutions for on-chip communication, since they avoid any form of routing and arbitration through the selection of disjoint carrier wavelengths for initiator-target pairs \cite{4211948,5948588,6096345}. 


Switching fabrics in a wavelength-routed \ac{ONoC} are generally implemented with microring resonators \cite{SiliconMicroringResonators}. These devices have a periodic transmittance characteristic, which means that they end up on resonance not only with one optical signal, but also with all those signals (if used) that are modulated on carrier wavelengths that are also resonant wavelengths of the microrings. 

This issue raises a misrouting problem: one optical signal (or a significant fraction of its power) heading to a specific destination may end up being coupled onto another optical path, leading to a different destination. However, this problem has not been consistently addressed so far in \ac{ONoC} literature, since the emphasis has been mainly on making the case for on-chip optical communication. As a result, wavelength-routed \ac{ONoC} topologies are typically not refined with implementation details, but rather assessed by means of high-level power macromodels. The ultimate implication is that physical parameters such as microring resonator radii and carrier wavelengths are not selected, but simply addressed by means of symbolic assignments. Hence, the misrouting concern (in this paper explicitly addressed as a routing fault) is left in the background.

The unmistakable evidence of this trend is given by the fact that whenever research teams come up with actual photonic integrated circuits of wavelength-routed structures, the misrouting concern arises.
For instance, in \cite{4815479} a $4\times 4$ optical crossbar using wavelength routing is fabricated and tested. Since designers did not give too much importance to parameter selection during the design phase, they ended up choosing resonant peaks for their microring resonators that were not properly spaced throughout the available bandwidth. As a result, when injecting optical power on specific lightpaths, they detected  significant power on unintended output ports of the device as well (an effect named {\em optical crosstalk}). Once deployed in a real system, their refined implementation may result in a misrouting fault and/or in error-prone communications, from the functional viewpoint.  \citeN{4815479} consider this as a future optimization step of their work. Our research aims at bridging exactly this existing gap in wavelength-routed \ac{ONoC} literature.

In previous work \cite{ICCAD16}, we discussed the electronics and photonics design issues linked to the maximization of the 
parallelism in \acp{WRONoC}. As explained in that paper, the optimal design was found using \ac{ASP}, 
a technology still not very well known in that
research area. In this work, we take for granted the electronics and photonics issues and focus on the
computational issues related to this hard optimization problem.
We detail the \ac{ASP} program used to solve the problem, and experimentally compare its performance with a \ac{MILP} model.
Another related problem was solved in \cite{INOC17} through a \ac{MILP} formulation. In this paper, we address the same problem in another logic language, namely \ac{CLP(FD)}, and show that the \ac{CLP(FD)} formulation is competitive with \ac{MILP} and that it is easier to modify.

In the next section, we describe the two problems addressed in this paper.
After some preliminaries (Section~\ref{sec:preliminaries}),
we formalize the problem of maximizing the parallelism in a \ac{WRONoC} (Section~\ref{sec:cop}),
then we describe the ASP program that solves such problem (Section~\ref{sec:asp_maxparall}).
We then motivate the second problem, namely the uniform spreading of the selected resonances, and propose a \ac{CLP(FD)} solution (Section~\ref{sec:spacing}).
We show through experimental results (Section~\ref{sec:experiments}) that the proposed logic programming approaches have good performance
with respect to mathematical programming formulations, and, finally, we conclude.

\section{Problem description}
\label{sec:problem_description}

In \acp{WRONoC}, \Ndisp\ senders communicate with \Ndisp\ receivers;
each source-destination pair is associated with 
an optical channel using a specific wavelength for the optical carrier:
the information originating from one sender is routed toward the correct
receiver depending on the used carrier wavelength.
In the same way, each receiver is able to receive communications from each of the \Ndisp\ senders, distinguishing the correct sender
through the wavelength of the carrier.
For simplicity, instead of {\em wavelength of the carrier} we will often use just {\em wavelength} or {\em carrier}.
Sender $\Sender_1$ uses disjoint 
wavelengths $\lamb_1$ to $\lamb_\Ndisp$ to communicate with
receivers $\Receiver_1$ to $\Receiver_\Ndisp$, respectively;
at the same time, receiver $\Receiver_1$ receives optical packets from senders $\Sender_1$ to $\Sender_\Ndisp$ on different wavelengths $\lamb_1$ to $\lamb_\Ndisp$. More in general:
\begin{itemize}
\item each sender uses different wavelengths to communicate with the different receivers; 
\item each receiver receives information from different senders using different wavelengths.
\end{itemize}
Instead of using a new set of wavelengths, sender $\Sender_2$ reuses the same wavelengths used by $\Sender_1$.

The communication flows of a \ac{WRONoC} topology can thus be abstracted by means of a Latin Square, that is
a matrix $\Ndisp\times\Ndisp$ containing \Ndisp\ values such that each row and each column contains \Ndisp\ values.
Each matrix value indicates the wavelength of the optical carrier that implements the communication between a specific sender-receiver pair.

\begin{figure}
\scalebox{.6} 
{
\begin{pspicture}(0,-1.445)(21.727188,1.485)
\usefont{T1}{ptm}{m}{n}
\rput(2.9026563,0.89){\large $\lambda_1$}
\psframe[linewidth=0.06,dimen=outer](3.5853126,1.355)(1.9853125,0.555)
\psline[linewidth=0.04cm,arrowsize=0.05291667cm 2.0,arrowlength=1.4,arrowinset=0.4]{->}(1.3853126,1.155)(1.9853125,1.155)
\psline[linewidth=0.04cm,arrowsize=0.05291667cm 2.0,arrowlength=1.4,arrowinset=0.4]{->}(1.3853126,0.755)(1.9853125,0.755)
\psline[linewidth=0.04cm,arrowsize=0.05291667cm 2.0,arrowlength=1.4,arrowinset=0.4]{->}(3.5853126,1.155)(10.785313,1.155)
\usefont{T1}{ptm}{m}{n}
\rput(11.702656,0.89){\large $\lambda_3$}
\psframe[linewidth=0.06,dimen=outer](12.385312,1.355)(10.785313,0.555)
\usefont{T1}{ptm}{m}{n}
\rput(7.302656,-0.11){\large $\lambda_2$}
\psframe[linewidth=0.06,dimen=outer](7.9853125,0.355)(6.3853126,-0.445)
\psline[linewidth=0.04cm,arrowsize=0.05291667cm 2.0,arrowlength=1.4,arrowinset=0.4]{->}(12.385312,1.155)(17.385313,1.155)
\usefont{T1}{ptm}{m}{n}
\rput(0.7359375,1.265){(1,2,3,4)A}
\usefont{T1}{ptm}{m}{n}
\rput(0.718125,0.865){(1,2,3,4)B}
\usefont{T1}{ptm}{m}{n}
\rput(7.158125,1.305){(1)A (2,3,4)B}
\usefont{T1}{ptm}{m}{n}
\rput(5.1759377,0.465){(1)B (2,3,4)A}
\usefont{T1}{ptm}{m}{n}
\rput(9.011094,0.465){(1)C (2)A (3,4)D}
\usefont{T1}{ptm}{m}{n}
\rput(19.028906,1.265){(1)C (2)A (3)B (4)D : $\Receiver_1$}
\usefont{T1}{ptm}{m}{n}
\rput(19.028906,0.265){(1)D (2)C (3)A (4)B : $\Receiver_2$}
\usefont{T1}{ptm}{m}{n}
\rput(2.9026563,-1.11){\large $\lambda_1$}
\psframe[linewidth=0.06,dimen=outer](3.5853126,-0.645)(1.9853125,-1.445)
\psline[linewidth=0.04cm,arrowsize=0.05291667cm 2.0,arrowlength=1.4,arrowinset=0.4]{->}(1.3853126,-0.845)(1.9853125,-0.845)
\psline[linewidth=0.04cm,arrowsize=0.05291667cm 2.0,arrowlength=1.4,arrowinset=0.4]{->}(1.3853126,-1.245)(1.9853125,-1.245)
\usefont{T1}{ptm}{m}{n}
\rput(11.702656,-1.11){\large $\lambda_3$}
\psframe[linewidth=0.06,dimen=outer](12.385312,-0.645)(10.785313,-1.445)
\psline[linewidth=0.04cm,arrowsize=0.05291667cm 2.0,arrowlength=1.4,arrowinset=0.4]{->}(12.385312,-1.245)(17.385313,-1.245)
\psline[linewidth=0.04cm,arrowsize=0.05291667cm 2.0,arrowlength=1.4,arrowinset=0.4]{->}(3.5853126,-1.245)(10.785313,-1.245)
\usefont{T1}{ptm}{m}{n}
\rput(0.7217187,-0.735){(1,2,3,4)C}
\usefont{T1}{ptm}{m}{n}
\rput(0.73109376,-1.135){(1,2,3,4)D}
\usefont{T1}{ptm}{m}{n}
\rput(6.971719,-1.055){(1)D (2,3,4)C}
\usefont{T1}{ptm}{m}{n}
\rput(5.1717186,-0.735){(1)D (2,3,4)C}
\usefont{T1}{ptm}{m}{n}
\rput(9.025937,-0.735){(1)B (2)D (3,4)A}
\usefont{T1}{ptm}{m}{n}
\rput(16.102655,-0.11){\large $\lambda_4$}
\psframe[linewidth=0.06,dimen=outer](16.785313,0.355)(15.185312,-0.445)
\usefont{T1}{ptm}{m}{n}
\rput(14.211094,0.665){(1)A (2,4)B (3)D}
\usefont{T1}{ptm}{m}{n}
\rput(14.225938,-0.735){(1)D (2,4)C (3)A }
\usefont{T1}{ptm}{m}{n}
\rput(19.028906,-1.135){(1)B 2(D) (3)C (4)A : $\Receiver_4$}
\usefont{T1}{ptm}{m}{n}
\rput(19.028906,-0.135){(1)A (2)B (3)D (4)C : $\Receiver_3$}
\psline[linewidth=0.04,arrowsize=0.05291667cm 2.0,arrowlength=1.4,arrowinset=0.4]{->}(7.9853125,-0.245)(10.185312,-0.245)(10.185312,-0.845)(10.785313,-0.845)
\psline[linewidth=0.04,arrowsize=0.05291667cm 2.0,arrowlength=1.4,arrowinset=0.4]{->}(3.5853126,-0.845)(4.1853123,-0.845)(4.1853123,-0.245)(6.3853126,-0.245)
\psline[linewidth=0.04,arrowsize=0.05291667cm 2.0,arrowlength=1.4,arrowinset=0.4]{->}(3.5853126,0.755)(4.1853123,0.755)(4.1853123,0.155)(6.3853126,0.155)
\psline[linewidth=0.04,arrowsize=0.05291667cm 2.0,arrowlength=1.4,arrowinset=0.4]{->}(12.385312,0.755)(12.985312,0.755)(12.985312,0.155)(15.185312,0.155)
\psline[linewidth=0.04,arrowsize=0.05291667cm 2.0,arrowlength=1.4,arrowinset=0.4]{->}(12.385312,-0.845)(12.985312,-0.845)(12.985312,-0.245)(15.185312,-0.245)
\psline[linewidth=0.04,arrowsize=0.05291667cm 2.0,arrowlength=1.4,arrowinset=0.4]{->}(16.785313,0.155)(17.185312,0.155)(17.385313,0.155)
\psline[linewidth=0.04,arrowsize=0.05291667cm 2.0,arrowlength=1.4,arrowinset=0.4]{->}(7.9853125,0.155)(10.185312,0.155)(10.185312,0.755)(10.785313,0.755)
\psline[linewidth=0.04cm,arrowsize=0.05291667cm 2.0,arrowlength=1.4,arrowinset=0.4]{->}(16.785313,-0.245)(17.385313,-0.245)
\end{pspicture} 
}
\caption{A \ac{WRONoC} topology connecting 4 senders (named A, B, C and D) to 4 receivers ($\Receiver_1$ to $\Receiver_4$) and using 4 carrier wavelengths (named $\lambda_1$ to $\lambda_4$).
Numbers refer to communication channels, e.g., $(1,2)$A means the communication channels consisting of $\lambda_1$ and $\lambda_2$ originating from sender A.\label{fig:lambdarouter}}
\end{figure}
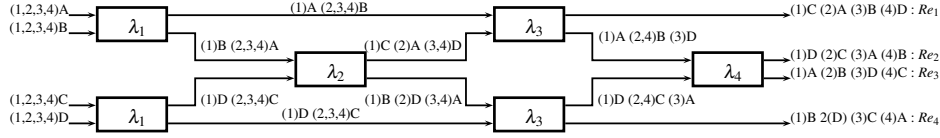

The routing is done through optical devices called \acp{PS}; typical \acp{PS} 
have two input 
and two output ports and have a base resonance wavelength.
They consist of two micro-rings, and the base resonance wavelength depends on the radius of the rings.
If the signal in the first input port resonates with the \ac{PS}, then it is deviated toward the first output; otherwise
it is passed to the second output port. The second input is treated symmetrically.
A number of such devices build up a \ac{WRONoC}, and various topologies have been proposed to ensure the correct routing of the information.
Figure~\ref{fig:lambdarouter} shows one of such topologies, connecting four senders (A, B, C, and D) to four receivers ($\Receiver_1$ to $\Receiver_4$), and using
four wavelengths ($\lambda_1,\dots,\lambda_4$). For example, if sender A uses wavelength $\lambda_1$, the signal resonates with the first \ac{PS}
and exits from the first output port; here it is sent to a \ac{PS} that resonates with $\lambda_3$ and is sent to its second output port.
It then enters the first input port of a \ac{PS} resonating with $\lambda_4$ and is then sent to its second output port.
Note that each of the four receivers can distinguish the origin of the information through the used wavelength; e.g., when
$\Receiver_3$
senses a signal of wavelength $\lambda_1$, the sender must have been A.

It can be observed that each \ac{PS} can resonate not only to its base wavelength, but also to a number of other
harmonic wavelengths; Table~\ref{tab:RL} is an example of a very small instance showing the set
$\SetPossLambda_r =\{\lambda_{r,1},\lambda_{r,2},\dots\}$ of
resonance wavelengths for different values of radii.
This effect can be exploited to increase the communication parallelism, as a  sender-receiver pair could communicate
not only through the
base wavelength but also using some of the harmonics.
In such a case, the communication channel between two endpoints consists of two or more carriers, with different wavelengths,
resonances of the same radius.
However, it might be the case that the $i$-th harmonic of one \ac{PS} could be equal (or too close) to the $j$-th
harmonic of another one: in such a case the laser beam would be incorrectly deviated in the \ac{WRONoC} topology, and a so-called {\em misrouting} or {\em routing 
fault} would occur.

\begin{figure}
\scalebox{.6} 
{
\begin{pspicture}(0,-2.72)(15.235937,2.32)
\definecolor{color3068}{rgb}{0.0,0.06274509803921569,1.0}
\definecolor{color3129}{rgb}{0.00784313725490196,0.7686274509803922,0.1411764705882353}
\definecolor{color3193}{rgb}{0.00392156862745098,0.00392156862745098,0.00392156862745098}
\psarc[linewidth=0.04,linecolor=red](1.0,1.7){1.0}{-90.0}{0.0}
\psline[linewidth=0.04cm,linecolor=red](3.0,0.7)(4.0,0.7)
\psarc[linewidth=0.04,linecolor=red](4.0,1.7){1.0}{-90.0}{0.0}
\psarc[linewidth=0.04,linecolor=red](3.0,1.7){1.0}{-180.0}{-90.0}
\psline[linewidth=0.04cm,linecolor=red](6.0,0.7)(7.0,0.7)
\psarc[linewidth=0.04,linecolor=red](7.0,1.7){1.0}{-90.0}{0.0}
\psarc[linewidth=0.04,linecolor=red](6.0,1.7){1.0}{-180.0}{-90.0}
\psline[linewidth=0.04cm,linecolor=red](9.0,0.7)(10.0,0.7)
\psarc[linewidth=0.04,linecolor=red](10.0,1.7){1.0}{-90.0}{0.0}
\psarc[linewidth=0.04,linecolor=red](9.0,1.7){1.0}{-180.0}{-90.0}
\psline[linewidth=0.04cm,linecolor=color3068](3.4,-0.9)(4.0,-0.9)
\psarc[linewidth=0.04,linecolor=color3068](4.0,0.1){1.0}{-90.0}{0.0}
\psarc[linewidth=0.04,linecolor=color3068](3.4,0.1){1.0}{-180.0}{-90.0}
\psline[linewidth=0.04cm,linecolor=color3068](1.0,-0.9)(1.4,-0.9)
\psarc[linewidth=0.04,linecolor=color3068](1.4,0.1){1.0}{-90.0}{0.0}
\psline[linewidth=0.04cm,linecolor=color3068](6.0,-0.9)(6.6,-0.9)
\psarc[linewidth=0.04,linecolor=color3068](6.6,0.1){1.0}{-90.0}{0.0}
\psarc[linewidth=0.04,linecolor=color3068](6.0,0.1){1.0}{-180.0}{-90.0}
\psline[linewidth=0.04cm,linecolor=color3068](8.6,-0.9)(9.2,-0.9)
\psarc[linewidth=0.04,linecolor=color3068](9.2,0.1){1.0}{-90.0}{0.0}
\psarc[linewidth=0.04,linecolor=color3068](8.6,0.1){1.0}{-180.0}{-90.0}
\psline[linewidth=0.04cm,linecolor=color3068](11.2,-0.9)(11.8,-0.9)
\psarc[linewidth=0.04,linecolor=color3068](11.8,0.1){1.0}{-90.0}{0.0}
\psarc[linewidth=0.04,linecolor=color3068](11.2,0.1){1.0}{-180.0}{-90.0}
\psarc[linewidth=0.04,linecolor=color3068](13.8,0.1){1.0}{-180.0}{-90.0}
\psline[linewidth=0.04cm,linecolor=red](12.0,0.7)(13.0,0.7)
\psarc[linewidth=0.04,linecolor=red](13.0,1.7){1.0}{-90.0}{-33.690067}
\psarc[linewidth=0.04,linecolor=red](12.0,1.7){1.0}{-180.0}{-90.0}
\psarc[linewidth=0.04,linecolor=color3129](14.4,-1.5){1.0}{180.0}{236.30994}
\psline[linewidth=0.04cm,linecolor=color3129](1.0,-2.5)(1.4,-2.5)
\psarc[linewidth=0.04,linecolor=color3129](1.4,-1.5){1.0}{-90.0}{0.0}
\psline[linewidth=0.04cm,linecolor=color3129](3.4,-2.5)(3.6,-2.5)
\psarc[linewidth=0.04,linecolor=color3129](3.6,-1.5){1.0}{-90.0}{0.0}
\psarc[linewidth=0.04,linecolor=color3129](3.4,-1.5){1.0}{-180.0}{-90.0}
\psline[linewidth=0.04cm,linecolor=color3129](5.6,-2.5)(5.8,-2.5)
\psarc[linewidth=0.04,linecolor=color3129](5.8,-1.5){1.0}{-90.0}{0.0}
\psarc[linewidth=0.04,linecolor=color3129](5.6,-1.5){1.0}{-180.0}{-90.0}
\psline[linewidth=0.04cm,linecolor=color3129](7.8,-2.5)(8.0,-2.5)
\psarc[linewidth=0.04,linecolor=color3129](8.0,-1.5){1.0}{-90.0}{0.0}
\psarc[linewidth=0.04,linecolor=color3129](7.8,-1.5){1.0}{-180.0}{-90.0}
\psline[linewidth=0.04cm,linecolor=color3129](10.0,-2.5)(10.2,-2.5)
\psarc[linewidth=0.04,linecolor=color3129](10.2,-1.5){1.0}{-90.0}{0.0}
\psarc[linewidth=0.04,linecolor=color3129](10.0,-1.5){1.0}{-180.0}{-90.0}
\psline[linewidth=0.04cm,linecolor=color3129](12.2,-2.5)(12.4,-2.5)
\psarc[linewidth=0.04,linecolor=color3129](12.4,-1.5){1.0}{-90.0}{0.0}
\psarc[linewidth=0.04,linecolor=color3129](12.2,-1.5){1.0}{-180.0}{-90.0}
\psline[linewidth=0.04cm,linecolor=color3193](1.0,2.3)(1.0,-2.7)
\psline[linewidth=0.04cm,linecolor=color3193,arrowsize=0.05291667cm 2.0,arrowlength=1.4,arrowinset=0.4]{->}(0.6,-2.7)(13.8,-2.7)
\usefont{T1}{ptm}{m}{n}
\rput(2.1853125,2.04){\Large $\lambda_{11}$}
\usefont{T1}{ptm}{m}{n}
\rput(5.1853123,2.04){\Large $\lambda_{12}$}
\usefont{T1}{ptm}{m}{n}
\rput(8.185312,2.04){\Large $\lambda_{13}$}
\usefont{T1}{ptm}{m}{n}
\rput(11.185312,2.04){\Large $\lambda_{14}$}
\usefont{T1}{ptm}{m}{n}
\rput(2.5853126,0.44){\Large $\lambda_{21}$}
\usefont{T1}{ptm}{m}{n}
\rput(5.1853123,0.44){\Large $\lambda_{22}$}
\usefont{T1}{ptm}{m}{n}
\rput(7.7853127,0.44){\Large $\lambda_{23}$}
\usefont{T1}{ptm}{m}{n}
\rput(10.385312,0.44){\Large $\lambda_{24}$}
\usefont{T1}{ptm}{m}{n}
\rput(12.985312,0.44){\Large $\lambda_{25}$}
\usefont{T1}{ptm}{m}{n}
\rput(2.5853126,-1.16){\Large $\lambda_{31}$}
\usefont{T1}{ptm}{m}{n}
\rput(4.7853127,-1.16){\Large $\lambda_{32}$}
\usefont{T1}{ptm}{m}{n}
\rput(6.9853125,-1.16){\Large $\lambda_{33}$}
\usefont{T1}{ptm}{m}{n}
\rput(9.185312,-1.16){\Large $\lambda_{34}$}
\usefont{T1}{ptm}{m}{n}
\rput(11.385312,-1.16){\Large $\lambda_{35}$}
\usefont{T1}{ptm}{m}{n}
\rput(13.585313,-1.16){\Large $\lambda_{36}$}
\usefont{T1}{ptm}{m}{n}
\rput(0.7353125,1.44){\Large $r_1$}
\usefont{T1}{ptm}{m}{n}
\rput(0.7353125,-0.36){\Large $r_2$}
\usefont{T1}{ptm}{m}{n}
\rput(0.7353125,-2.16){\Large $r_3$}
\end{pspicture} 
}
\caption{\label{fig:ondine} An example of available spectrum, with three available radii $\SetPossRadius=\{r_1,r_2,r_3\}$
	and a set of resonating wavelengths for each radius}
\end{figure}
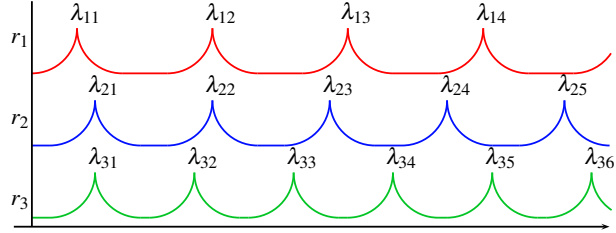

In Figure~\ref{fig:ondine} three possible radius values are available $\SetPossRadius=\{r_1,r_2,r_3\}$;
for each radius $r_i$, there is a set of resonating wavelengths $\{r_{i,1},r_{i,2},\dots\}$ that can be selected as carriers.
Suppose that $\Ndisp=2$; this means that 2 radii must be selected (out of the 3 available).
Note that $\lambda_{21}=\lambda_{31}$; this means that if both $r_2$ and $r_3$ are selected, wavelength $\lambda_{21}$ cannot
be selected as carrier, because it would be incorrectly routed, since it also resonates with radius $r_3$.
The same holds also for $\lambda_{12}=\lambda_{22}$.
Also, the wavelength $\lambda_{14}$ is very close to $\lambda_{35}$; in real settings there exist always imprecisions in the
fabrication process, so it is not advisable to select wavelengths that are too close: a minimum distance $\Dl$ should separate
any two selected wavelengths.

One possible solution would be to select $r_1$ and $r_2$; in such a case, three wavelengths can be selected for each radius without routing faults: 
in fact
for $r_1$ the set of wavelengths $\{\lambda_{11},\lambda_{13},\lambda_{14}\}$ can be selected,
while for $r_2$ any three wavelengths can be selected out of the four that do not conflict with $r_1$: $\lambda_{21}$, $\lambda_{23}$, $\lambda_{24}$, and $\lambda_{25}$. The obtained parallelism is 3.

\begin{table}[tbh]
\centering
 \caption{\RLStable\ with radii varying from 5 to $8\mu$m  
 \label{tab:RL}}
 \begin{tabular}{c|@{}c|@{}c|@{}lllllll@{}}
 \hline
   $r$ & $R_r$   & $|\{\lambda_{r,j}\}|$ & $\lambda_{r,1}$ [nm]&$\lambda_{r,2}$ [nm] &$\lambda_{r,3}$ [nm] &$\lambda_{r,4}$ [nm] &$\lambda_{r,5}$ [nm] &$\lambda_{r,6}$ [nm] &$\lambda_{r,7}$ [nm]\\
   \hline
  $1$ & $5\mu$m & 5 & $1496.4$& $1521.3$& $1547.1$& $1573.8$& $1601.4$\\
  $2$ & $6\mu$m & 6 & $1500.5$& $1521.3$& $1542.7$& $1564.8$& $1587.5$& $1610.8$\\
  $3$ & $7\mu$m & 6 & $1503.4$& $1521.3$& $1539.6$& $1558.4$& $1577.7$& $1597.4$ \\
  $4$ & $8\mu$m & 7 & $1505.6$& $1521.3$& $1537.3$& $1553.7$& $1570.4$& $1587.5$& $1604.9$ \\
\hline
 \end{tabular}
\end{table}

It is then important to select \Ndisp\ different radii, taken from the set of available radii \SetPossRadius, 
and for each selected radius $r$ select \Nlambdas\ resonating wavelengths (taken from the set $\SetPossLambda_r$ 
of harmonics of the radius $r$)
such that
each sender-receiver pair can use \Nlambdas\ wavelengths (harmonics of the same radius)
while avoiding routing faults;
the objective is maximizing the number \Nlambdas.
This problem was solved in \cite{ICCAD16} through an \ac{ASP} formulation, that was only cited in that paper.
In this paper, instead, we detail the \ac{ASP} program in Section~\ref{sec:asp_maxparall}.

After finding the maximum parallelism $\Nlambdas$ obtainable, one has to 
choose a suitable solution amongst the (possibly, many) solutions providing the same optimal value of parallelism.
In \cite{INOC17}, it was found that the wavelengths found when solving the first problem could be unevenly spread in the available spectrum.
This introduced a second problem: given \Ndisp\ and \Nlambdas, find \Ndisp\ radii values and $\Ndisp\times\Nlambdas$ wavelengths (\Nlambdas\ per 
radius) such that the selected wavelengths are as evenly spread as possible.
Such problem was solved in \cite{INOC17} with a \ac{MILP} formulation.
In this work, we address the same problem in another logic programming language, namely \ac{CLP(FD)}.
We show that the \ac{CLP(FD)} program is competitive in terms of performance with the \ac{MILP} approach.
Moreover, we found that a different formulation is more adherent to the \ac{WRONoC} design problem,
and that
the CLP(FD) program can be easily modified to account for the revised formulation. The MILP approach, instead,
must be subject to major rewriting in order to tackle this revised formulation.

The complete solution process consists of two phases: in the first, the maximum obtainable parallelism is obtained through an \ac{ASP} program. The optimal value of parallelism is then provided to the second phase: a \ac{CLP(FD)} program that,
 given a target value of parallelism, computes a set of wavelengths that 
1) achieve the given
parallelism level and 2) are  
 as equally spaced as possible in the available spectrum.

\section{Preliminaries}
\label{sec:preliminaries}

\subsection{Answer Set Programming}
\label{sec:asp}

Answer Set Programming (ASP) is a class of logic programming languages that rely on the stable model semantics \cite{StableModels},
also known as answer set semantics.
We assume a basic familiarity with logic programming and its syntax; for an introduction 
the reader can refer to \cite{Lloyd}.
A logic program consists of a set of rules 
$a \mbox{~:-~} l_1, l_2, \dots, l_n$
where $a$ is an atom (also called the {\em head} of the rule), and the $l_i$ are literals (called the {\em body} of the rule).

Literals and rules containing no variables are called {\em ground}. We denote as $gr(r)$ all
possible instantiations of the rule $r$ of the program $\Pi$,
on the basis of ground facts of the program.
The {\em ground istantiation} of $\Pi$ consists of
all ground instances of rules in $\Pi$,
i.e., $gr(\Pi)=\bigcup_{r \in \Pi} gr(r)$.
For any set $M$ of atoms from $\Pi$, let $\Pi_M$ be the program obtained from $\Pi$ by deleting $(i)$ each rule that has a negative literal $\neg B$ in its body with $B\in M$ and $(ii)$ all negative literals in the bodies of the remaining rules. Since $\Pi_M$ is negation free, it has a unique minimal Herbrand model. If this model coincides with $M$, then $M$ is a Stable Model of $\Pi$ \cite{StableModels}.


Among the dialects of \ac{ASP}, 
we use the language of the grounder Gringo \cite{Gringo},
that extends the basic logic programming syntax with a number of features.


%

\noindent {\em Counting} 
\cite{WeightConstraints}.
If $a_1,a_2,a_3,\dots$ are atoms, and $l$ and $u$ are integers,
the aggregate
$l  \{ a_1, a_2, a_3, \dots \}  u$
is true for every set $S$ of atoms including from $l$ to $u$ members of $\{ a_1, a_2, a_3, \dots \}$,
i.e., $l \leq | \{ a_i \in S \} | \leq u$. Trivial bounds can be omitted.

\noindent {\em Summation}.
If $a_1, a_2, a_3, \dots$ are atoms and $v_1, v_2, v_3,\dots$ are integers, the
aggregate
$l  {\bf\sharp sum} [ a_1 = v_1, a_2 = v_2, a_3 = v_3, \dots ]  u$
is true for every set S of atoms such that the sum of $v_i$ over included members
$a_i$ of $\{ a_1, a_2, a_3, \dots \}$ is in the interval $[l,u]$:
$l \leq \sum_{i:a_i\in S} v_i \leq u.$

Usually, ASP solvers \cite{WeightConstraints,ASSAT,Cmodels,DLV_ToCL,potassco} work in two stages.
In the first, called {\em grounding}, the program is converted into an equivalent ground program. 
The second stage is devoted to looking for stable models (answer sets) of the ground program.

\subsection{Constraint Logic Programming on Finite Domains}

\acf{CLP} is a class of logic programming languages \cite{JM94} that extends Prolog with the notion of {\em constraints}.
Each language of the CLP class is identified with a {\em sort};
one of the most popular is CLP(FD), on the sort of Finite Domains.
CLP(FD) is particularly suited to solve \acp{CSP}.
A \ac{CSP} consists of a set of decision variables, each ranging on a finite domain,
and subject to a set of relations called {\em constraints}.
A solution to the CSP is an assignment of values taken from the domains to the respective variables,
such that all the constraints are satisfied.

A \ac{COP} is a \ac{CSP} with an additional objective function, that must be maximized or minimized.
A solution of a COP is optimal if it satisfies all the constraints and, amongst the solutions of the CSP,
it maximizes (or minimizes) the objective function.

\section{Maximizing parallelism}
\label{sec:cop}
We now give a formalization of the problem of finding the maximum parallelism.
A set $\SetPossRadius$ of possible radius values is given.
For each $r \in \SetPossRadius$, a set $\SetPossLambda_r = \{\lamb_{rj}\}$ of resonance wavelengths is also given.
Two wavelengths $\lamb_{ri}$, $\lamb_{sj}$ are in conflict if $|\lamb_{ri}- \lamb_{sj}| \leq \Dl$ for a given $\Dl \geq 0$.

The core decisions concern which resonances should be selected for each radius.
To model this decision we use the boolean variable $x_{rj}\in\{0,1\}$ to state whether the resonance wavelength $\lamb_{rj}$ is selected for radius $r$.
The problem can be formalized as the following \ac{COP}:
\begin{align}
 P(s)&=\max: \min_{r\in 1..|\SetPossRadius|}\{q_r\mid q_r>0\}\;\;\text{s.t.} & \label{cop:of}\\
 q_r &=\sum_{\lamb_{rj} \in \SetPossLambda_{r}} x_{rj}& \forall\ r \in 1..|\SetPossRadius| \label{cop:qi}\\
s_r &= 
\begin{cases}
0 & q_r=0\\
1 & q_r>0
\end{cases} & \forall\ r \in 1..|\SetPossRadius|  \label{cop:si}\\
\sum_{r \in 1..|\SetPossRadius|} s_r & =\Ndisp & \label{cop:stot}
%
\\
x_{rj}=1 &\Rightarrow s_{r'} = 0 &
\begin{array}{r@{}}
\forall r, r' \in 1..|\SetPossRadius| \forall j \in 1..|\SetPossLambda_r| \\\mbox{ s.t. } \exists i \in 1..|\SetPossLambda_{r'}| \land |\lambda_{rj}-\lamb_{r'i}|\leq \Dl
\end{array}
  \label{cop:routfault}
\end{align}
$q_r$ represents the number of selected resonances for radius $r$.
The objective function~(\ref{cop:of}) maximizes the parallelism in the selected radius with the least parallelism,
since the global network parallelism is bounded by the channel with lowest parallelism.
In practice, we maximize the minimum parallelism that can be sustained by all of the wavelength channels.
Constraints~(\ref{cop:qi}) define the number $q_r$ of selected elements in row $r$.
Constraints~(\ref{cop:si}) define whether the radius $r$ is selected ($s_r=1$) or not ($s_r=0$).
Constraint~(\ref{cop:stot}) imposes to select exactly $\Ndisp$ radii.
Finally, Constraints (\ref{cop:routfault}) prevent routing faults;
they are imposed for each $\lamb_{rj}$ and $r'\neq r$ such that $\lambda_{rj}$ is conflicting with some resonance 
$\lambda_{r'i}$ in radius
$r'$. 



Consider, for example, the instance in Table~\ref{tab:RL}, suppose that $\Ndisp=3$,
i.e., three radii must be selected, and $\Dl=0$, i.e., two wavelengths are in conflict only if they are identical.
One solution is to select radii 2, 3, and 4, i.e., $s_1=0$ and $s_2=s_3=s_4=1$ (satisfying constraint~\ref{cop:stot}).
Notice that in Table~\ref{tab:RL} $\lambda_{r,2}=1521.3$ for all values of $r$. From constraint~\ref{cop:routfault},
selecting such wavelength for some radius (e.g., for radius 2, i.e. $x_{22}=1$) means that all other radii must not be selected: contradiction. Thus clearly $x_{r,2}=0$ for all radii $r$.
Also, $\lambda_{2,5}=\lambda_{4,6}$, so by constraint~(\ref{cop:routfault}),
they cannot be selected, since both radii 2 and 4 are selected.
All other wavelengths can be selected; i.e. $x_{2,1}=x_{3,1}=x_{4,1}=x_{3,3}=\dots =x_{4,7}=1$ is a possible assignment.
We have that $q_1=0$, $q_2=4$, $q_3=5$ and $q_4=5$. The minimum of the not-null $q_i$ is $q_2=4$, that is also the value of the
objective function for this assignment.

\section{An ASP program to compute maximum WRONoC parallelism}
\label{sec:asp_maxparall}

The \ac{ASP} program takes as input an instance provided with 
facts
%
\begin{lstlisting}
lambda($R$,$L_{min}$,$L_{nominal}$,$L_{max}$)
\end{lstlisting}
expressing the fact that the radius $R$ resonates at the wavelength $L_{nominal}$; due to variations in temperature and other uncertainties, the 
actual wavelength might change, with a maximum deviation \Dl, i.e., in the range $[L_{min},L_{max}] = [L_{nominal}-\Dl,L_{nominal}+\Dl]$.

Predicate \pred{radius}{1} is true for the available radii (the elements of the set \SetPossRadius), while \pred{lambda}{2} is true for the available wavelengths for each radius (elements of the set $\SetPossLambda_R$):
\begin{lstlisting}
lambda($R$,$L$) :- lambda($R$,_,$L$,_).
radius($R$) :- lambda($R$,_).
\end{lstlisting}
From the set of available wavelengths, some are chosen as transmission carriers. Predicate 
{\tt sL}$(r,j)$
is true if the wavelength $j$ is chosen for the radius $r$, i.e., iff $x_{r,j}=1$ in the \ac{COP} of Eq.~(\ref{cop:of}-\ref{cop:routfault}):
\begin{lstlisting}
{ sL($R$,$L$) : lambda($R$,$L$) }.
\end{lstlisting}
The set of chosen radii is then given by:
\begin{lstlisting}
sR($R$) :- sL($R$,_).
\end{lstlisting}
{\tt sR($r$)} is true iff $s_r=1$ in the \ac{COP} of Eq.~(\ref{cop:of}-\ref{cop:routfault}).
The number of chosen radii must be equal to the number \Ndisp\ of devices that need to communicate:
\begin{equation}
\mbox{\tt :- not }\Ndisp \leq \mbox{\tt \{ sR($R$): radius($R$) \} } \leq \Ndisp.
\label{eq:asp_number_radii}
\end{equation}
%
%
In order to avoid routing faults (constraint~(\ref{cop:routfault})), we define a conflict relation.
Two wavelengths $L1$ and $L2$ are in conflict if they are selected for different radii
and the intervals $[L^1_{min},L^1_{max}]$ and $[L^2_{min},L^2_{max}]$ have non-empty intersection.
\begin{lstlisting}
conflict($R1$,$R2$,$L1$,$L2$):- lambda($R1$,$L^1_{min}$,$L1$,$L^1_{max}$), $R1$!=$R2$,
                       lambda($R2$,$L^2_{min}$,$L2$,$L^2_{max}$), $L1<L2$, $L^1_{max}\geq L^2_{min}.$
conflict($R1$,$R2$,$L1$,$L2$):- lambda($R1$,$L^1_{min}$,$L1$,$L^1_{max}$), $R1$!=$R2$, 
                       lambda($R2$,$L^2_{min}$,$L2$,$L^2_{max}$), $L1$>$L2$, $L^2_{max} \geq L^1_{min}.$
conflict($R1$,$R2$,$L$,$L$):- lambda($R1$,$L^1_{min}$,$L$,$L^1_{max}$), $R1$!=$R2$,
                     lambda($R2$,$L^2_{min}$,$L$,$L^2_{max}$), 
\end{lstlisting}
Also, it might be the case that two wavelengths for the same radius are in conflict
\begin{lstlisting}
conflict($R$,$L1$,$L2$):- lambda($R$,$L^1_{min}$,$L1$,$L^1_{max}$), 
	lambda($R$,$L^2_{min}$,$L2$,$L^2_{max}$), $L1$<$L2$, $L^1_{max}$>=$L^2_{min}$.
\end{lstlisting}
Note that the {\tt conflict} predicate depends only on the input data, and not on the wavelengths that must be chosen as carriers.
The truth of the {\tt conflict} atoms in the answer set is decided in the grounding phase, and does not 
require a search during the computation of the answer set.

If wavelength $L1$ of radius $R1$ is in conflict with some wavelength of radius $R2$, then $L1$ and $R2$ cannot be both selected;
if two wavelengths are in conflict within the same radius, they cannot be selected:
\begin{lstlisting}
:-conflict($R1$,$R2$,$L1$,$L2$),radius($R1$),radius($R2$),sL($R1$,$L1$),sR($R2$).
:-conflict($R$,$L1$,$L2$), sL($R$,$L1$), sL($R$,$L2$), $ L1$<$L2$.
\end{lstlisting}
Finally, the objective is to maximize the number of wavelengths selected for each radius.
Predicate \pred{countR}{2} provides the number of selected resonances for each radius,
and corresponds to constraint~(\ref{cop:qi}):
\begin{lstlisting}
countR($R$,$Qr$) :- radius($R$), $Qr$>0, $Qr$=#count{ 1,$L$ : sL($R$,$L$) }.
\end{lstlisting}
Predicate \pred{bp}{1} provides the minimum number of resonances that have been selected varying the radius;
the objective is maximizing such value, as in Eq~(\ref{cop:of}):
\begin{lstlisting}
#maximize{ $P$ : bp($P$) }.
\end{lstlisting}

Predicate \pred{bp}{1} could be implemented following the definition (Eq.~\ref{cop:of}), i.e.:
\begin{lstlisting}
bp($P$):- $P$=#min{ $Qr$:countR($R$,$Qr$) }, $P$>0.
\end{lstlisting}
%
however, a more efficient version is using chaining and an auxiliary
predicate:\footnote{We thank one of the anonymous reviewers for suggesting this improved formulation.}
\begin{lstlisting}
auxbp(N):- countR(_,N).
auxbp(N+1) :- auxbp(N), N < F, maxF(F).
bp(P):- auxbp(P), not auxbp(P-1).
\end{lstlisting}
where {\tt maxF} computes the maximum number of wavelengths that might be selected,
and that can be calculated during grounding.

\section{Spacing the selected resonances}
\label{sec:spacing}

As will be shown in the experimental results (Section~\ref{sec:experiments}), the ASP program in Section~\ref{sec:asp_maxparall} was very efficient in computing the maximum parallelism.
On the other hand, after analyzing the provided solutions, it was found that often the selected wavelengths were unevenly spread in the available spectrum. 
Since, due to imprecisions in the fabrication process, the actual wavelengths might be different from the computed ones, it might be the case that
two selected wavelengths become too close in the actual device, and the two wavelengths might be confused raising a routing fault.
As often done in the electronic component industry, after fabrication each device is checked, and if it is not working properly it is discarded.

A second-level optimization could then be performed in order to select, amongst the possibly many resonances that provide the same optimal 
parallelism, those ones that are more evenly spread in the available spectrum, with the idea that maximizing the distance between 
selected wavelengths can reduce the likelihood that the actual wavelengths are too close, and, consequently, that the device has to be discarded.

The \ac{ASP} program in Section~\ref{sec:asp_maxparall} was then modified to
take as input the parallelism to be achieved, and to have as objective to uniformly spread the selected resonances.
The performances, however, were not satisfactory, and a complex \ac{MILP} model, based on network flow, was devised \cite{INOC17}.
Another logic programming based approach was developed in \ac{CLP(FD)}; we describe it in next section.

%
%
%
%

\subsection{A \ac{CLP(FD)} approach to the problem of spacing selected resonances}
\label{sec:clp}

As already said, in this second optimization phase, we have as input a value \Nlambdas\
of parallelism to be achieved. The objective is to select \Ndisp\ values of radii and $\Ndisp\times\Nlambdas$
resonance wavelengths (\Nlambdas\ for each radius) such that the selected wavelengths are as equally spread
in the available spectrum as possible.

In the \ac{CLP} program, we focused on modeling the problem with fewer variables than the \ac{MILP} and ASP formulations.
In \ac{MILP} the problem is modeled with one variable for each pair $(r,\lamb)$ stating that resonance $\lamb$ is selected for radius $r$.
Similarly, in \ac{ASP} there is predicate {\tt sL}$(r,\lamb)$ that is true if $\lamb$ is selected for radius $r$.
In the \ac{CLP} program, we have \Ndisp\ variables $R_1,\dots,R_\Ndisp$ that range over the set \SetPossRadius\ of possible radii;
each of the $R_i$ represents one chosen value of radius.

A common rule of thumb to have efficient CLP(FD) programs is to employ the so-called {\em global constraints} \cite{alldifferent},
i.e., constraints that involve a large number of variables and for which powerful propagation algorithms have been designed in the past.
The idea is that using global constraints, the propagation can exploit more global information (opposed to the local information used
in arc-consistency propagation and its variants) because each constraint has visibility of many variables at the same time.

Clearly, all the radii must be different, so we have
$$\alldifferent([R_1,\dots,R_\Ndisp])$$
where $\alldifferent$ \cite{alldifferent} imposes that all variables take different values.

The selected resonances are represented through an $\Ndisp\times\Nlambdas$ matrix $\Mat$;
each element $\Mat_{ij}$ ranges over the set of available wavelengths, and represents the $j$-th wavelength selected for radius $R_i$.

Each of the variables in the $i$-th row of matrix $\Mat$ is linked to the radius variable $R_i$;
for each $i$, $\Mat_{ij}$ should be a resonance wavelength of radius $R_i$.
This can be imposed through 
a {\tt table} constraint \cite{tableConstraintNengFa}.
The {\tt table} constraint is useful to define new constraints by listing the set of available tuples;
in our case it lists the set of pairs $(R,L)$ for which $R$ is a radius and $L$ one of its corresponding resonance wavelengths.

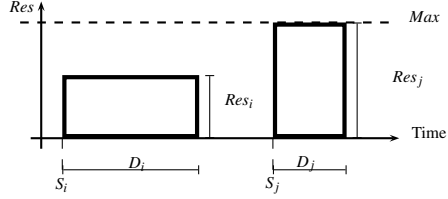
\begin{figure}
\scalebox{.7} 
{
\begin{pspicture}(0,-1.908125)(8.602813,1.908125)
\psline[linewidth=0.04cm,arrowsize=0.05291667cm 2.0,arrowlength=1.4,arrowinset=0.4]{->}(0.5809375,-0.7903125)(7.5809374,-0.7903125)
\psline[linewidth=0.04cm,arrowsize=0.05291667cm 2.0,arrowlength=1.4,arrowinset=0.4]{->}(0.7809375,-0.9903125)(0.7809375,1.8096875)
\psframe[linewidth=0.08,dimen=outer](3.7809374,0.4096875)(1.1809375,-0.7903125)
\psline[linewidth=0.04cm,linestyle=dashed,dash=0.16cm 0.16cm](0.3809375,1.4096875)(7.3809376,1.4096875)
\usefont{T1}{ptm}{m}{n}
\rput(8.062344,1.5196875){$Max$}
\usefont{T1}{ptm}{m}{n}
\rput(0.41234374,1.7196875){$Res$}
\usefont{T1}{ptm}{m}{n}
\rput(8.150312,-0.6803125){Time}
\usefont{T1}{ptm}{m}{n}
\rput(1.1723437,-1.6803125){$S_i$}
\psline[linewidth=0.02cm,linestyle=dashed,dash=0.16cm 0.16cm](1.1809375,-0.7903125)(1.1809375,-1.3903126)
\psline[linewidth=0.02cm,tbarsize=0.07055555cm 5.0]{|-|}(1.1809375,-1.3903126)(3.7809374,-1.3903126)
\usefont{T1}{ptm}{m}{n}
\rput(2.6023438,-1.2803125){$D_i$}
\psline[linewidth=0.02cm,tbarsize=0.07055555cm 5.0]{|-|}(3.9809375,0.4096875)(3.9809375,-0.7903125)
\usefont{T1}{ptm}{m}{n}
\rput(4.5423436,-0.0803125){$Res_i$}
\psframe[linewidth=0.08,dimen=outer](6.5809374,1.4096875)(5.1809373,-0.7903125)
\usefont{T1}{ptm}{m}{n}
\rput(5.182344,-1.6803125){$S_j$}
\psline[linewidth=0.02cm,linestyle=dashed,dash=0.16cm 0.16cm](5.1809373,-0.7903125)(5.1809373,-1.3903126)
\psline[linewidth=0.02cm,tbarsize=0.07055555cm 5.0]{|-|}(5.1809373,-1.3903126)(6.5809374,-1.3903126)
\usefont{T1}{ptm}{m}{n}
\rput(5.8123436,-1.2803125){$D_j$}
\psline[linewidth=0.02cm,tbarsize=0.07055555cm 5.0]{|-|}(6.7809377,1.4096875)(6.7809377,-0.7903125)
\usefont{T1}{ptm}{m}{n}
\rput(7.7523437,0.3196875){$Res_j$}
\end{pspicture} 
}
\caption{Example of cumulative constraint with two tasks \label{fig:example_cumulative}}
\end{figure}

Constraint Logic Programming is particularly effective at solving scheduling problems, mainly due to the effectiveness of the $cumulative$ constraint.
The $cumulative$ constraint considers a set of tasks, each described with a start time, a duration and a resource consumption,
and it ensures that in each time the sum of the resources consumed by the scheduled tasks does not exceed a given limit $Max$.
Let $S$ be the list of start times, $D$ that of durations and $Res$ that of resource consumptions, 
$$cumulative(S,D,Res,Max)$$
is true if (see Figure~\ref{fig:example_cumulative})
$$\forall t \qquad
\sum_{i: S_i \leq t \leq S_i+D_i} Res_i \leq Max.$$
The three lists $S$, $D$ and $Res$ can contain variables with domains or constant values,
and the constraint removes, through constraint propagation, inconsistent values.
In the particular case in which $\forall i, Res_i=1$ and $Max=1$, the $cumulative$ constraint imposes that the tasks should not overlap in time.

In the problem of maximally spreading wavelengths, we model the selected wavelengths as tasks of a scheduling problem.
Each of the $\Mat_{ij}$ elements of the $\Mat$ matrix is considered as the start time of a task, with a total of $\Ndisp\times\Nlambdas$ tasks.
All the tasks have the same duration: one variable $Dist$ models the duration of all the tasks.
If we now impose\footnote{where we indicate with $[X]_n$ the list $[X,X,X,...]$ containing $n$ times element $X$.}
\begin{equation}
cumulative([\Mat_{ij}|i\in 1..\Ndisp,j\in 1..\Nlambdas],[Dist]_\Ndisp,[1]_\Ndisp,1)
\label{eq:cumulative_spacing1}
\end{equation}
this constraint imposes that all wavelengths do not overlap, and that they are spaced of at least $Dist$ units.
The objective will be to find the maximum possible value for variable $Dist$ that does not cause any conflict.


To model conflicts between resonances, we recall that
each resonance wavelength for a chosen radius $R_k$ must be different from all the wavelengths $\Mat_{ij}$ selected for another radius $R_i$.
We first explain how to model the relation between a radius $R_i$ and the set of resonance wavelengths, then we provide
a set of global constraints to model conflicting wavelengths.

The relation between a radius $R_r$ and the corresponding resonances $\SetPossLambda_{R_r}=\{\lambda_{r,1},\lambda_{r,2},\dots\}$ is imposed through 
an {\tt element} constraint \cite{elementConstraint}.
The $element(I,L,X)$ constraint ensures that the $I$-th element of the list $L$ has value $X$.
We represent the set $\SetPossLambda_r$ as a list of constrained variables $[\lambda_{r,1},\lambda_{r,2},\dots]$; the length of the list is the number of resonances in the radius with the maximum number of resonances $\MaxLambdasPerRadius = \max_k |\SetPossLambda_k|$.
%
%
The $i$-th element of the list, $\lambda_{r,i}$, is subject to the constraint
$$element(R_r,\RLStable^i,\lambda_{r,i})$$
where $\RLStable^i$ is the $i$-th column of Table~\ref{tab:RL}.
To account for the different number of resonances in different radii, the list is filled with dummy values.

Since the list of resonance wavelengths consists of different wavelengths, in order to model conflicts between the selected resonances for one radius and the resonance wavelengths for other radii
one might impose
\begin{equation}
\forall i \in 1..\Ndisp, \forall k \in 1..\Ndisp, i\neq k, \quad \alldifferent([\Mat_{ij}|j \in 1..\Nlambdas]\cup \SetPossLambda_{R_k})
\label{eq:n2_alldifferent}
\end{equation}
that are 
$\Ndisp(\Ndisp-1)$
$\alldifferent$ constraints, each containing 
$\Nlambdas+ \MaxLambdasPerRadius$ 
variables.
However, one might notice as well that all the elements in the $\Mat$ matrix are different, meaning that
instead of (\ref{eq:n2_alldifferent}) one can impose
\begin{equation}
\forall k \in 1..\Ndisp, \quad \alldifferent([\Mat_{ij}|i \in 1..\Ndisp, i\neq k, j \in 1..\Nlambdas]\cup \SetPossLambda_{R_k})
\label{eq:n_alldifferent}
\end{equation}
that are $\Ndisp$ constraints each containing 
$(\Ndisp-1)\Nlambdas+ \MaxLambdasPerRadius$
variables.


Please, note the symbols: each radius $r$ has a number of resonance wavelengths, the $j$-th is named $\lambda_{rj}$.
Out of the $\lambda_{rj}$, some are selected as carriers: the $i$-th wavelength selected for radius $r$ is named $\Mat_{ri}$.

Finally, the objective is maximizing variable $Dist$, that is a lower bound to the minimal distance between selected wavelengths.

\subsubsection{Breaking Symmetries}
The problem contains a number of symmetries:
\begin{itemize}
\item The order in which the resonance wavelengths appear in one of the rows of the matrix $L$ is not important: given a solution, another solution can be obtained by swapping two elements. More importantly, swapping two elements in an assignment that is not a solution, provides another non-solution.
\item Swapping the order of two radii (both in the list of radii and as rows of the $\Mat$ matrix) provides an equivalent solution.
\end{itemize}
Removing symmetries is considered important to speedup the search.
We tried several strategies, and the best was the following:
\begin{itemize}
\item the rows of the $\Mat$ matrix are sorted in ascending order. This could be done imposing $\Mat_{ij}<\Mat_{i,j+1}$, but since all wavelengths must be at least $Dist$ units apart, the following constraint gives stronger propagation:
$$\forall i \in 1..\Ndisp, \forall j \in 1..\Nlambdas-1 \quad \Mat_{ij} + Dist \leq \Mat_{i,j+1}$$
\item the first column of the matrix is sorted in ascending order:
$$\forall i \in 1..\Ndisp-1 \quad \Mat_{i,1} + Dist \leq \Mat_{i+1,1}$$
\end{itemize}

\subsubsection{Objective function}

As previously said, the objective is to maximize the value assigned to variable $Dist$, that represents the minimum distance between two
selected resonances.

Adding known bounds of the objective function can strengthen the propagation.
Clearly, the maximum possible value for $Dist$ is obtained if all the selected wavelengths are equally spaced.
As $\Ndisp\Nlambdas$ resonances are selected, the following bound holds:
$$(\Ndisp\Nlambdas-1)Dist \leq \left(\max_{i \in 1..\Ndisp, j \in 1..\Nlambdas} \Mat_{i,j}\right) - \left(\min_{i \in 1..\Ndisp, j \in 1..\Nlambdas} \Mat_{i,j}\right).$$
Given the symmetry breaking constraints, $\min_{i,j} \Mat_{i,j} = \Mat_{1,1}$,
while $\max_{i,j} \Mat_{i,j}$
is the maximum of the last column of the $\Mat$ matrix:
$ \max_{i} \Mat_{i,\Ndisp}$.

%

\subsection{A refined CLP(FD) approach}
\label{sec:refined_clp}

As will be shown in the experimental results (Section~\ref{sec:experiments}),
the CLP approach just shown  did not reach the performance of the \ac{MILP} program in \cite{INOC17}.
However, a closer look to the set of selected wavelengths (both in the MILP and in the CLP approaches) 
showed that a further refinement of the problem formulation was necessary.
In fact, in order to minimize the likelihood of routing faults, 
a selected resonance $\Mat_{r,i}$ should be as far as possible not only from the other selected resonances $\Mat_{s,j}$,
but also from all the resonance wavelengths of the selected radii ($\lambda_{R,i}$ for all the selected $R$ and all $i$), independently from the fact that they are also selected
as carriers or not.

Considering this effect, the \ac{MILP} model in \cite{INOC17} can  no longer be used 
and a major rewriting is required, because the problem can no longer be modeled as a constrained shortest path.

The \ac{CLP} program, instead, can be easily modified to account for this effect.
A first tentative would be to consider also the (non-selected) resonance wavelengths of selected radii
as tasks. The \alldifferent\ Constraints in Eq.~(\ref{eq:n_alldifferent}) can be rewritten as $cumulative$ constraints,
in which the tasks corresponding to selected wavelengths have duration $\Dist$, while those corresponding to non-selected wavelengths have
a very short duration (value 1nm is suitable in our instances):
$$\begin{array}{l@{}l@{}l}
\forall k \in 1..\Ndisp, \ \cumulative(&[\Mat_{ij}|i \in 1..\Ndisp, i\neq k, j \in 1..\Nlambdas]++\SetPossLambda_k,&
\\&[\Dist]_{(\Ndisp-1)\Nlambdas}++[1]_{\MaxLambdasPerRadius},&[1]_{(\Ndisp-1)\Nlambdas+\MaxLambdasPerRadius},1)
\end{array}
$$
where the symbol {\tt ++} stands for list concatenation.
However, with this approach each selected resonance would be at least \Dist\ units from the 
{\em following} resonance (either selected or non-selected), but no constraint prevents it to be very close to the {\em preceding} 
non-selected resonance.

A possible solution would be to represent selected resonances $\Mat_{ij}$
as tasks with start time $\Mat_{ij}-\frac{\Dist}{2}$ and duration $\Dist$, i.e., $\Mat_{ij}$ would be the center of the task instead of its start time.
This modification introduces a large overhead, due to the fact that the constraint associated with the summation operator propagates very poorly.

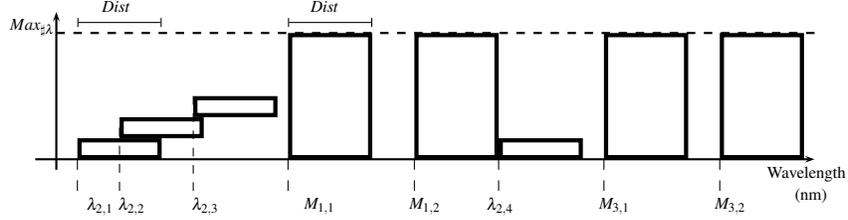
\begin{figure}
\scalebox{.7} 
{
\begin{pspicture}(0,-2.108125)(16.137188,2.108125)
\psline[linewidth=0.04cm,arrowsize=0.05291667cm 2.0,arrowlength=1.4,arrowinset=0.4]{->}(0.599375,-0.9903125)(15.399375,-0.9903125)
\psline[linewidth=0.04cm,arrowsize=0.05291667cm 2.0,arrowlength=1.4,arrowinset=0.4]{->}(0.999375,-1.1903125)(0.999375,1.8096875)
\psframe[linewidth=0.08,dimen=outer](2.999375,-0.5903125)(1.399375,-0.9903125)
\psline[linewidth=0.04cm,linestyle=dashed,dash=0.16cm 0.16cm](0.999375,1.4096875)(15.399375,1.4096875)
\usefont{T1}{ptm}{m}{n}
\rput(0.55078125,1.5196875){\MLPR}
\usefont{T1}{ptm}{m}{n}
\rput(1.8307812,-1.8803124){$\lambda_{2,1}$}
\psline[linewidth=0.02cm,linestyle=dashed,dash=0.16cm 0.16cm](1.399375,-0.9903125)(1.399375,-1.5903125)
\psline[linewidth=0.02cm,tbarsize=0.07055555cm 5.0]{|-|}(1.399375,1.6096874)(2.999375,1.6096874)
\usefont{T1}{ptm}{m}{n}
\rput(2.139375,1.9196875){\Dist}
\psframe[linewidth=0.08,dimen=outer](6.999375,1.4096875)(5.399375,-0.9903125)
\usefont{T1}{ptm}{m}{n}
\rput(6.000781,-1.8803124){$\Mat_{1,1}$}
\psline[linewidth=0.02cm,linestyle=dashed,dash=0.16cm 0.16cm](5.399375,-0.9903125)(5.399375,-1.5903125)
\psline[linewidth=0.02cm,tbarsize=0.07055555cm 5.0]{|-|}(5.399375,1.6096874)(6.999375,1.6096874)
\usefont{T1}{ptm}{m}{n}
\rput(6.1007814,1.9196875){$\Dist$}
\psframe[linewidth=0.08,dimen=outer](3.799375,-0.1903125)(2.199375,-0.5903125)
\usefont{T1}{ptm}{m}{n}
\rput(2.4307814,-1.8803124){$\lambda_{2,2}$}
\psline[linewidth=0.02cm,linestyle=dashed,dash=0.16cm 0.16cm](2.199375,-0.3903125)(2.199375,-1.5903125)
\psframe[linewidth=0.08,dimen=outer](5.199375,0.2096875)(3.599375,-0.1903125)
\usefont{T1}{ptm}{m}{n}
\rput(3.8307812,-1.8803124){$\lambda_{2,3}$}
\psline[linewidth=0.02cm,linestyle=dashed,dash=0.16cm 0.16cm](3.599375,0.2096875)(3.599375,-1.5903125)
\psframe[linewidth=0.08,dimen=outer](9.399375,1.4096875)(7.799375,-0.9903125)
\psframe[linewidth=0.08,dimen=outer](12.999375,1.4096875)(11.399375,-0.9903125)
\psframe[linewidth=0.08,dimen=outer](15.199375,1.4096875)(13.599375,-0.9903125)
\psframe[linewidth=0.08,dimen=outer](10.999375,-0.5903125)(9.399375,-0.9903125)
\usefont{T1}{ptm}{m}{n}
\rput(8.000781,-1.8803124){$\Mat_{1,2}$}
\psline[linewidth=0.02cm,linestyle=dashed,dash=0.16cm 0.16cm](7.799375,-0.9903125)(7.799375,-1.5903125)
\usefont{T1}{ptm}{m}{n}
\rput(11.600781,-1.8803124){$\Mat_{3,1}$}
\psline[linewidth=0.02cm,linestyle=dashed,dash=0.16cm 0.16cm](11.399375,-0.9903125)(11.399375,-1.5903125)
\usefont{T1}{ptm}{m}{n}
\rput(13.800781,-1.8803124){$\Mat_{3,2}$}
\psline[linewidth=0.02cm,linestyle=dashed,dash=0.16cm 0.16cm](13.599375,-0.9903125)(13.599375,-1.5903125)
\usefont{T1}{ptm}{m}{n}
\rput(9.430781,-1.8803124){$\lambda_{2,4}$}
\psline[linewidth=0.02cm,linestyle=dashed,dash=0.16cm 0.16cm](9.399375,-0.9903125)(9.399375,-1.5903125)
\usefont{T1}{ptm}{m}{n}
\rput(15.245937,-1.2803125){Wavelength}
\usefont{T1}{ptm}{m}{n}
\rput(15.334531,-1.6803125){(nm)}
\end{pspicture} 
}
\caption{\label{fig:cumulativeFinal} Example of \cumulative\ for spacing the selected resonances ($\Mat_{i,j}$)
at a minimum distance $\Dist$.
Non-selected resonances ($\lambda_{i,j}$) can be close to each other,
but they cannot be close to selected resonances. 
}
\end{figure}

A more effective CLP(FD) modeling is to introduce a duration \Dist\ also for non-selected resonances (of selected radii).
However, this would introduce a minimal distance also between two non-selected wavelengths, a constraint which is not required for \acp{WRONoC}, and 
would lead to sub-optimal solutions.
We decided to use the {\em resource} parameter of the \cumulative\ constraint to avoid the collision between tasks of non-selected resonances.
Each of the $\MaxLambdasPerRadius$ non-selected resonances is modelled as a task of duration \Dist\ and using 1 resource unit
(see Figure~\ref{fig:cumulativeFinal}).
The limit of resources is exactly
\MaxLambdasPerRadius,
so that tasks of non-selected resonances can overlap.
Each selected resonance is modeled as a task of duration \Dist\ and using all resources 
(\MaxLambdasPerRadius):
$$\begin{array}{l@{}l@{}l}
\forall k \in 1..\Ndisp, \ \cumulative(&[\Mat_{ij}|i \in 1..\Ndisp, i\neq k, j \in 1..\Nlambdas]++\SetPossLambda_k,&
\\&[\Dist]_{(\Ndisp-1)\Nlambdas+\MaxLambdasPerRadius},[\MaxLambdasPerRadius]_{(\Ndisp-1)\Nlambdas}++[1]_{\MaxLambdasPerRadius},\MaxLambdasPerRadius)
\end{array}
$$
In this way a task corresponding to a selected resonance cannot overlap neither with tasks of selected resonances, nor with those of non-selected resonances, and must be at least at \Dist\ distance from any other resonance of selected radii.

\section{Experimental results}
\label{sec:experiments}

In the experimental campaign in \cite{ICCAD16}, the focus was computing the maximum obtainable parallelism varying the
fabrication parameters, including the possible deviations of the laser wavelengths and the radius imprecisions during fabrication of the device.
In this work, instead, we report the timing results of the \ac{ASP} formulation and of a \ac{MILP} model.

We considered a set of radii ranging from 5nm to 30nm in steps of 0.25nm; this yields 104 possible radii.
In order to compute the corresponding resonance wavelengths, 
an Electromagnetic Model~\cite{ramini} computes the transmission responses; with the selected values of radii,
1850 resonances are obtained, with a number of resonances per radius ranging from 5 to 28.

We compare the \ac{ASP} program described in Section~\ref{sec:asp_maxparall} with a \ac{MILP} model that is a linearization
(with standard techniques) of the problem defined in Section~\ref{sec:cop}.
The employed ASP solver is clasp 4.5.4, and the MILP solver is Gurobi 7.0.1;
Gurobi was run through its Python interface.
All experiments were run on a computer with an Intel Core i7-3720QM CPU running at 2.60GHz, with 16GB RAM,
using Linux Mint 18.1 64-bit. All experiments were performed using only one core.
All the code and instances are available on the web.\footnote{http://www.ing.unife.it/en/research/research-1/information-technology/computer-science/artificial-intelligence-group/research-projects/wronoc/}

\begin{figure}[tbh]
\includegraphics[width=.495\textwidth]{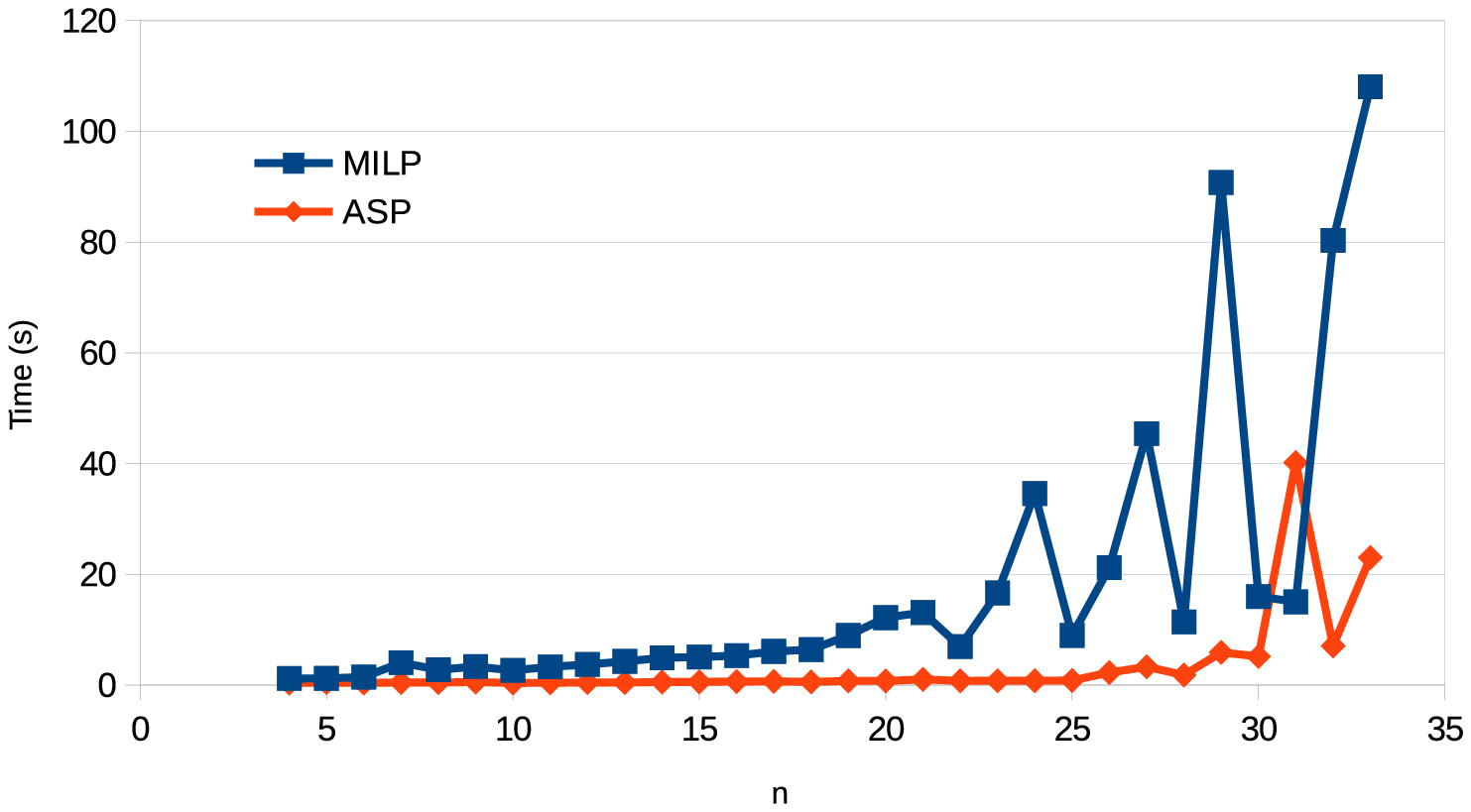}
\hfill
\includegraphics[width=.495\textwidth]{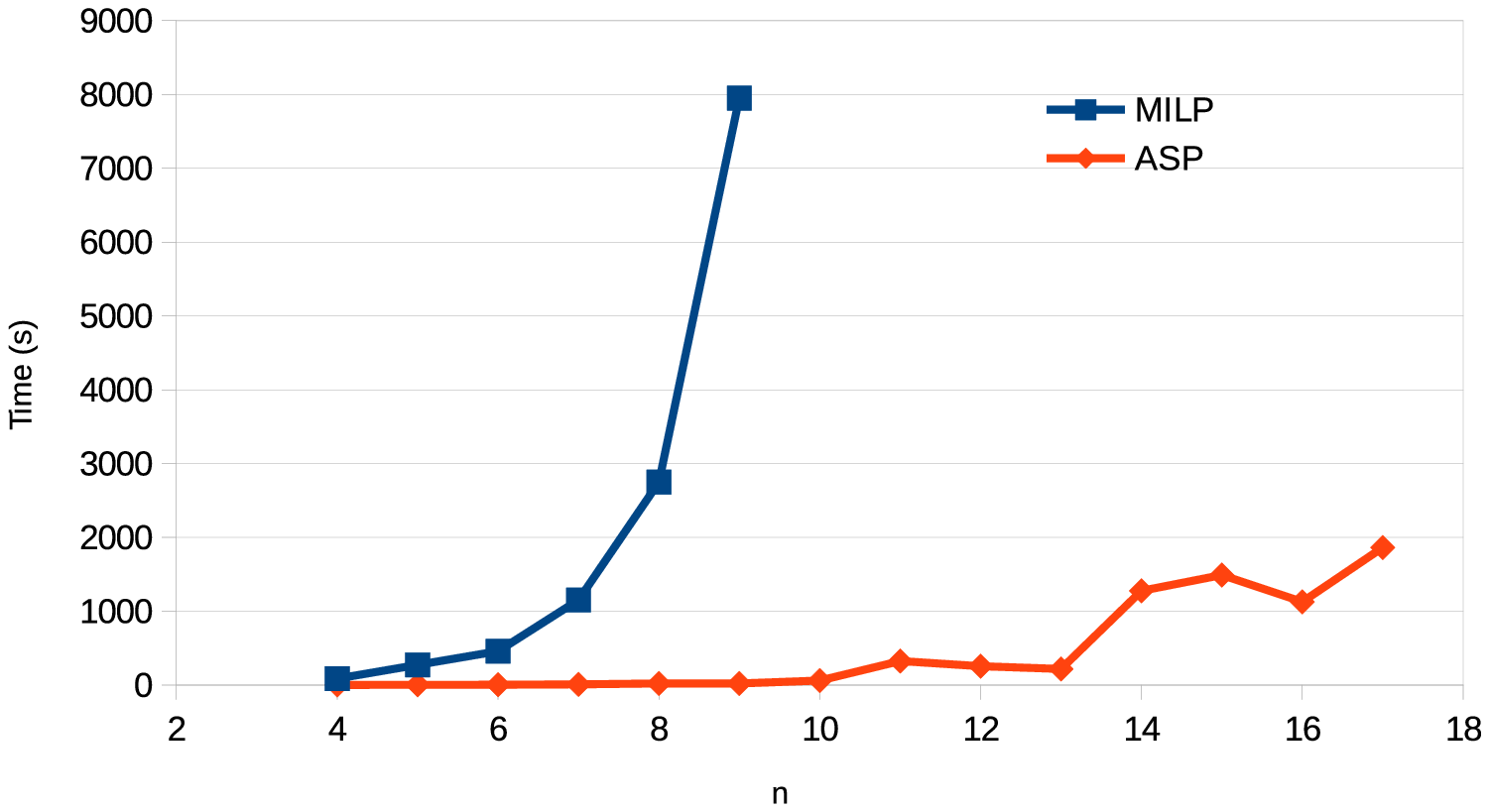}
\caption{\label{fig:milp_asp_DeltaLambda} Comparison of MILP and ASP running time when maximizing the minimum parallelism with $\Dl = 0$ (left) and $\Dl=1$nm (right)}
\end{figure}

The results are plotted in 
Figure~\ref{fig:milp_asp_DeltaLambda} for ideal lasers (left),
and for $\Dl=1$nm (right).
The ASP program has usually better performances than the MILP model,
and in particular in the non-ideal case, in which finding an assignment satisfying all constraints is more difficult,
while Gurobi seems more efficient in the case with less tight constraints, in which the difficulty is more driven by the
need to find an optimal solution.

%

\subsection{Maximally spreading resonances}

The second set of experiments assesses the performance of Logic Programming approaches in the problem of selecting carrier wavelengths maximally spread on the available spectrum.
We compare the performance of the \ac{CLP(FD)} program described in Section~\ref{sec:clp} with the \ac{MILP} flow model 
in the same instances considered in \cite{INOC17} and for which Gurobi did not run out of memory.

\begin{table}[bht]
\begin{tabular}{crrr}
$\Ndisp \times \Nlambdas$ & MILP & CLP(FD) & refined CLP(FD) \\
\hline
$4 \times 1$ & 508.68 & 14.45 & 10.41 \\
$8\times 1$ & 563.47 & 24.81 & 393.14 \\
$4 \times 4$ & 2973.77 & Time Out & 2303.85 \\
\hline
\end{tabular}
\caption{Comparison of MILP-Gurobi and CLP(FD)-\eclipse\ run time on the problem of maximizing the distance between wavelengths.
MILP and CLP(FD) maximize the distance only between wavelengths selected as carriers, while the {\em refined CLP(FD)} model
finds selected wavelengths at the maximum distance to any resonance wavelength of selected radii.
}
\end{table}

The experiments were run on the same computer given  earlier, using Gurobi 7.0.1 as MILP solver and \eclipse\ 6.1 \cite{eclipse} as CLP(FD) solver.
The time-out was 3600s.
While the MILP approach is very effective in the largest instance, the CLP(FD) program is more effective in the small instances.
On the other hand, the refined CLP(FD) program, that models more closely the requirements of the \ac{WRONoC} architecture,
is better than the MILP approach in all instances.
Note also that the MILP program \cite{INOC17} is very tailored toward solving the problem of maximally spreading the selected resonances
and has to be completely rewritten for modifications of the problem, such as adding further constraints or changing the objective function.
The Logic Programming approaches, instead, are more general and modifiable, as can be seen from the relatively small modifications required to extend the first CLP approach (Section~\ref{sec:clp}) to the refined CLP program (Section~\ref{sec:refined_clp}).

It is also worth noting that Gurobi is a commercial program, while both clingo and \eclipse\ are developed as open-source programs.


\section{Conclusions}

We presented two problems arising in the industry of opto-electronic components, in particular in \acf{WRONoC} design.
The first problem, published in \cite{ICCAD16} arose because in the electronic research the maximal communication parallelism obtainable with a 
\ac{WRONoC} was unknown.
The problem was solved with an \ac{ASP} program, that was mentioned, but not described in detail, in \cite{ICCAD16}.
We described the ASP program and compared experimentally its performance with a \acf{MILP} approach.

The second problem \cite{INOC17} comes from the observation that, once the maximum parallelism level is found, it is also of interest to
design the \ac{WRONoC} in the safest way, despite small variations that might occur in the fabrication process.
In order to maximize the probability that
the device is able to function correctly, the selected wavelengths used as carriers have to be as far as possible one from the other.
Such problem was approached in \cite{INOC17} through a MILP formulation.
In this work, we presented a \acf{CLP(FD)} program, showed that it has  performances competitive with the MILP approach and found that
it is easier to modify it to take into consideration further aspects in the \ac{WRONoC} design.

In both cases, Logic Programming approaches have proven to be competitive with mathematical programming technologies,
and that Logic Programming has promising techniques to address problems in the new area of \acf{WRONoC} design.

In future work, we plan to address the two described problems combining the best features of \ac{CLP} and \ac{ASP}; a number of Constraint Answer Set Programming solvers have been proposed and are natural candidates for this research direction
\cite{DBLP:journals/amai/MellarkodGZ08,IDP,DBLP:journals/tplp/DrescherW10,MINGO,DBLP:conf/padl/BalducciniL12,DBLP:conf/kr/LiuJN12,DBLP:conf/jelia/BartholomewL14,DBLP:conf/iclp/SusmanL16}.

\bibliographystyle{acmtrans}
\bibliography{wronoc,ICCAD,cilc12}

\label{lastpage}
\end{document}